%% file: main.tex
\pdfoutput=1

\documentclass[11pt]{article}
\usepackage[final]{acl}
\usepackage{natbib}
\usepackage{times}
\usepackage{latexsym}

\usepackage[T1]{fontenc}

\usepackage[utf8]{inputenc}
\usepackage{float}
\usepackage{listings}
\usepackage{microtype}

\usepackage{inconsolata}

\usepackage{graphicx}

\usepackage{booktabs}

\usepackage{ragged2e} 
\usepackage{multirow}

\usepackage{amsmath}
\usepackage{tikz}
\usepackage{titlesec}
\titleclass{\subsubsubsection}{straight}[\subsection]
\newcounter{subsubsubsection}[subsubsection]
\renewcommand\thesubsubsubsection{\thesubsubsection.\arabic{subsubsubsection}}
\titleformat{\subsubsubsection}
  {\normalfont\normalsize\bfseries}{\thesubsubsubsection}{1em}{}
\titlespacing*{\subsubsubsection}{0pt}{3.25ex plus 1ex minus .2ex}{1.5ex plus .2ex}
\title{ 
Analysing Austrian Parliamentary Speeches by Predicting Party Affiliations
}

 \author{
     \textbf{Sarah Sulollari},
     \textbf{Vanja Karan},
     \textbf{Benjamin Roth}
     \\
     \text{University of Vienna}
     \\
     \small{\textbf{Correspondence:} \href{mailto:a01200233@univie.ac.at}{\texttt{a01200233@unet.univie.ac.at}}}
 }

\begin{document}
\maketitle
\begin{abstract}
This paper examines speech characteristics of the Austrian National Council members (CMs). We introduce analysis methodology based on a fine-tuned BERT-Model which assigns to every speech a probability distribution over Austrian parties. These are then aggregated at the speaker and party level. We perform an extensive evaluation to ensure the quality of the generated distributions prior to further analysis. Instead of considering mis-assignments of predicted parties as a performance issue, we leverage them to derive insights about the CMs and political developments. 
Our findings show that parties that govern together become more similar in their speech characteristics, while, analogously, parties that are in opposition move further away from governing parties. Moreover, CMs who switch party usually adapt their speech characteristics only after the switch occurs. Finally, we publicly release a data set containing party affiliation predictions for the speeches of the Austrian National Council from 1996 to 2018.

\end{abstract}

\section{Introduction}

While political scientists have long analyzed party behavior through manual discourse analysis, NLP enables us to systematically quantify speech patterns, where misclassifications reveal how politicians adapt their rhetoric: aligning with coalition partners or distancing themselves in opposition.
We are trying to find answers to the following research questions: 
Q1 :Can party political affiliations be modeled using parliamentary speeches?
Q2: How are coalition and opposition relations reflected in speech behaviour?
Q3: Does the switching of parties affect CM speeches? 

Specifically, we explore the party-political dynamics of Austrian parliamentary speeches from 1996 to 2018.  These speeches are public and represent a core part of our democratic society. Because of the vast number of speeches, this dataset represents an ideal test bed for our methodology. Furthermore, enabling analysis of this data can help both researchers and engaged citizens explore and better understand the dynamics in Parliamentary discussions. Thereby, it can help build a bridge from the National Council directly to the public.  Moreover, learning more about language use within the Parliament is interesting from the perspective of social sciences as well as linguistics.
Finally, our methodology could be applied to similar data from other domains, fostering research in related fields which currently mostly rely on qualitative manual methods.

We start from the Parlspeech V2 dataset \cite{DVNL4OAKN2020}, which contains speeches from the Austrian parliament dating back to 1996. Its metadata allows for identifying which council member (CM) was speaking and with which party they were affiliated.

\begin{table}
	\centering
    \small
	\begin{tabular}{lc}
    \toprule
    \textbf{Party} & \textbf{Political Leaning} \\ \midrule
    SPÖ & centre left\\
    ÖVP & centre right\\
    FPÖ & right\\
    GRÜNE & centre left\\
    LIF & centre right \\
    BZÖ & right \\
    NEOS & centre right \\
    STRONACH & right \\
    PILZ & left\\ 
    \bottomrule
    \end{tabular}
	\caption{Rough estimation of political parties in data set}
	\label{tab:party_leanings}
\end{table}

Our analysis methodology is based on training a probabilistic text classifier at the level of the speech (or speech chunks). The predictions of this classifier on held-out data are then aggregated at the level of CM or party. We thus obtain probabilistic vector representations of both speakers and parties, which can be visualised in different ways. In addition, further insights can be obtained by using prediction failures \footnote{Cases where the most probable predicted party drastically differs from the party officially listed in the metadata.} to guide the focus of qualitative analysis. Overall our approach can be seen as a two-step process: classifier training (cf.~Sections 4 and 5) and model output analysis (cf.~Section 6). We use the approach to investigate language use and developments in the Austrian Parliament. 

The contribution of this work is threefold. First, we develop and evaluate a methodology for deriving and analysing aggregated probability distributions over parties for each speech. Second, we use this methodology to produce a dataset of such probabilities for data from the Austrian National Council over a period of 22 years. Third, using the methodology, we conduct several analyses on the generated data, showcasing its usefulness for deriving meaningful insights.\footnote{The code and dataset will be made publicly available.}

\section{Related Work}
The introduction of BERT \cite{devlin2019bertpretrainingdeepbidirectional} and subsequent evaluations against alternative classifiers \cite{Garrido_Merchan_2023} have demonstrated the effectiveness of modern text classification approaches. While generative LLMs have been explored for tasks like elite polarization detection in the European Parliament \cite{iakovlev2025elitepolarizationeuropeanparliamentary}, recent comparisons show that fine-tuned BERT models often outperform prompted LLMs in text classification, particularly when contextual knowledge is required \cite{vajjala2025textclassificationllmera,zhang-etal-2025-bert-like}.

Early work by \newcite{gerrish2011predicting} developed predictive models linking legislative sentiment to legislative text to predict voting patterns. In the domain of political discourse, \newcite{reinig-etal-2024-politics} identified rhetorical differences between government and opposition through speech act analysis, while \newcite{patz2025analyzing} emphasized the need for specialized embeddings to capture political speech nuances. \newcite{navarretta2024government} used a pre-trained Danish BERT model to classify speeches by parties in government, opposition, or “other” roles, achieving F1-scores of 0.64 (3-class) and 0.77 (binary), and \newcite{abercrombie2022policy} identified policy preferences and speech sentiment in UK parliamentary debates using transformer-based models.

Other work has applied these techniques to different tasks. The GREEK-BERT model was fine-tuned for party prediction in both binary and multiclass tasks \cite{10345868}. In the Australian Parliament, stance classification compared BERT and methods based on word-counts. Surprisingly, the latter outperformed BERT in prediction performance and computational efficiency while offering greater transparency \cite{repec:spr:jcsosc:v:8:y:2025:i:2:d:10.1007_s42001-025-00366-y}. \newcite{doi:10.1086/730737} evaluated transformer models, highlighting the need for balanced training data—a particular challenge for small parties with limited speeches. \newcite{reinig-etal-2024-politics} also annotated speeches with speech acts using both support vector machines and fine-tuned BERT, finding that parties adopt different rhetorical strategies in opposition versus government, aligning with our findings.

The ParlSpeech V2 dataset \cite{DVNL4OAKN2020}, containing full-text corpora of 6.3 million parliamentary speeches from nine democracies, serves as the foundation for this paper. \newcite{skubic2024parliamentary} provide a literature review on parliamentary discourse research in political science, showing how parliamentary corpora can be integrated into political science research. 

In contrast to prior work, we use fine-tuned BERT to predict probabilistic party affiliations for each speech, aggregating these to analyze linguistic proximity between parties and politicians over time. Unlike previous studies, we leverage misclassifications to reveal political dynamics, such as coalition convergence and opposition divergence. Additionally, we publicly release a dataset of party affiliation predictions for Austrian parliamentary speeches from 1996 to 2018.
\section{Data}

The data set used in this paper is the ParlSpeech V2 data set \cite{DVNL4OAKN2020}, containing speeches of nine European Parliaments, including the Austrian National Council. 
The Austrian data includes speeches from 1,051 sittings from  January 1996 until December 2018 . Excluding the speeches held by the President of the Parliament and the Scribes the 88219 speeches are distributed between the nine parties.

\begin{figure}
    \centering
    \includegraphics[width=0.75\linewidth]{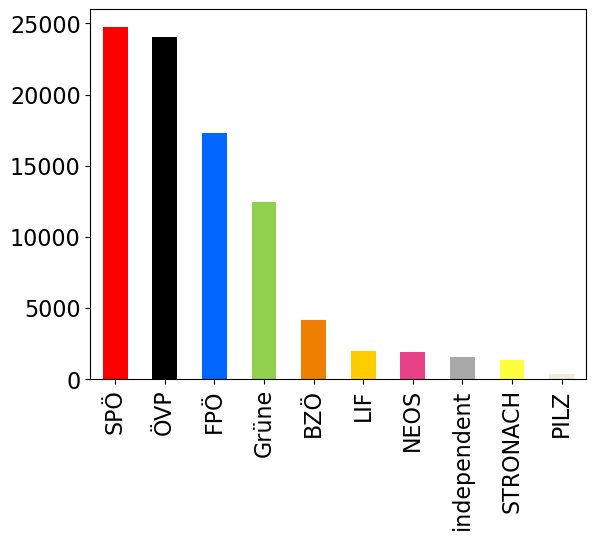}
    \caption{Distribution of Speeches across all parties}
    \label{fig:ch3_2}
\end{figure}

When looking at the speeches per party in Figure~\ref{fig:ch3_2}, it is important to consider that the parties have varying amounts of time and seats in Parliament.
Parties that were in Parliament since 1996 (SPÖ, ÖVP and FPÖ) have the most speeches followed by GRÜNE, which was in Parliament from 1996, apart from the XXVI legislative period (9 November 2017 to 22 October 2019), when they were not in Parliament.
In this period, PILZ was in Parliament consisting of some former GRÜNE CMs as well as one former member of SPÖ.
STRONACH only stayed in Parliament for one legislative period (XXV) but lost so many members to other parties or as independents, that by the end of the period they were no longer a Parliamentary Group.
Both NEOS and LIF only took part in two legislative periods.
Additionally, the number of seats each party had in Parliament differs greatly.
Three top parties, SPÖ, ÖVP and FPÖ were in government for some of the time covered in the data.

\subsection{Speakers}
\begin{figure}
		\centering
		\includegraphics[width=0.75\linewidth]{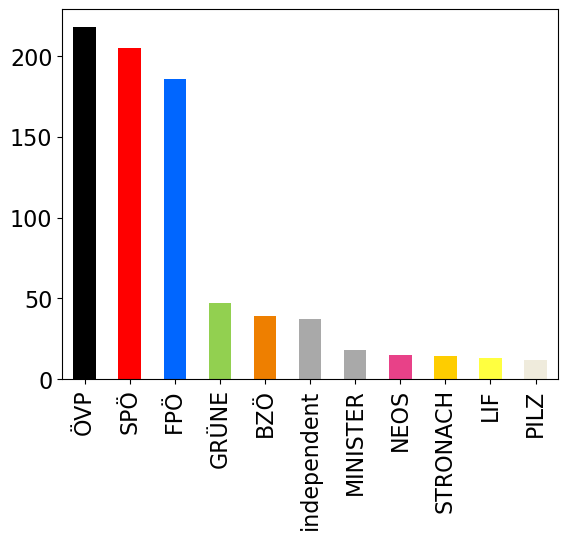}
		\caption{Distribution of speakers across parties.} 
		\label{fig:ch3_mps}
\end{figure}

In total the data set includes 708 speakers, of which 691 are CMs.
They are distributed amongst the parties as seen in Figure ~\ref{fig:ch3_mps}. 
In total, 55 CMs changed their party affiliation, including the CMs that became independent majority members of the National Council.  
The highest number of CMs left FPÖ and the fewest left LIF. 
Most CMs left a party and became independent.
The party that was most joined by other CMs was the BZÖ.

\begin{figure*}
		\centering
        \includegraphics[width=0.7\textwidth]{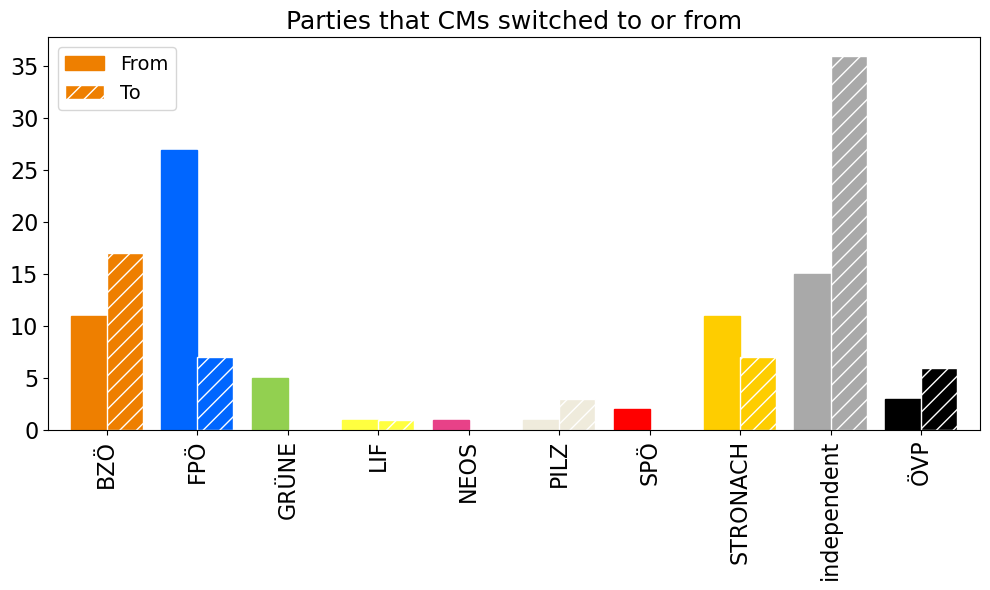}
		\caption{Number of CMs joining or leaving a party.}
		\label{fig:ch3_join_mps}
\end{figure*}

Figure ~\ref{fig:ch3_join_mps} compares the number of CMs that joined a party and left a party. 
Here we can clearly see that some parties have little to no movement while others are very flexible. 

\section{Classification Methodology}
\input{classificationmodelshort}

\subsection{Training details}

\paragraph{Folds} Because the aim is to collect a held-out prediction of the party for every speech, we use a 5 fold cross-validation setup. The folds are generated at the level of the CM \footnote{Speeches of a single CM are either all train or all test.} and we use stratified sampling to ensure class distribution in all folds is similar to the overall class distribution (Figure ~\ref{fig:ch3_mps}).

\paragraph{Architecture}

We use the The HuggingFace Transformers library~\cite{wolf-etal-2020-transformers} to fine-tune the \texttt{bert-base-german-cased} \href{https://huggingface.co/google-bert/bert-base-german-cased}{pretrained Model}~\cite{bert-base-german-cased}. On top of it is a classification head with a softmax activation which provides a probabilistic output for each of the nine parties.  

\paragraph{Context length}
We found that it is inevitable that some of the speeches will be too long to fit into the context length of the model. We resolve this by dividing speeches that are too long into chunks of 256 tokens, applying the classifier to each chunk separately and aggregating the results back to speech level by taking the mean. 

\paragraph{Parameters} We minimized cross entropy loss using a batch size of 20, learning rate 2e-5 and training for two epochs. In the final layer we apply a dropout of 10\% for regularisation. In preliminary experiments we tried slightly varying these parameters on a smaller test sample, but there were no significant differences in performance. 

\section{Classifier Evaluation}
While the performance on speech level is limited, we will show the classifier does perform well when analysed in detail and aggregated on speaker level. 

\subsection{Raw model scores}
For this evaluation we used the output probabilities directly, giving us a distribution over parties for each speech. If the speech had several chunks the corresponding distributions are aggregated(by taking a mean). For evaluation purposes the highest probability party was selected as the prediction for that speech.

The scores are of limited quality by standard text classification metrics.
FPÖ and ÖVP received the highest scores with roughly 60\% F1-score, followed by SPÖ and GRÜNE.
LIF, BZÖ and NEOS do not perform well with scores between 0.2 and 0.4.
STRONwACH and PILZ barely manage an F1-Score above 0.1.
\begin{table}
	\centering
    \small
	\begin{tabular}{lcccc}
    \toprule
    \textbf{Party} & \textbf{Precision} & \textbf{Recall} & \textbf{F1} & \textbf{Support} \\ \midrule
    SPÖ & 0.53 & 0.57 & 0.55 & 12109 \\ 
    ÖVP & 0.57 & 0.59 & 0.58 & 11477 \\ 
    FPÖ & 0.49 & 0.58 & 0.54 & 9736 \\ 
    GRÜNE & 0.60 & 0.45 & 0.51 & 7856 \\
    LIF & 0.43 & 0.31 & 0.35 & 1430
    \\
    BZÖ & 0.33 & 0.31 & 0.32 & 2389 \\ 
    NEOS & 0.38 & 0.29 & 0.33 & 1059 \\ 
    STRONACH & 0.14 & 0.10 & 0.11 & 761 \\ 
    PILZ & 0.09 & 0.07 & 0.07 & 190 \\ 
    \bottomrule
    \end{tabular}
	\caption{Average scores per speech aggregated per party}
	\label{tab:ch5_avrg_Scores}
\end{table}

For further evaluation, all speeches with fewer than 30 tokens were excluded, as these speeches usually contained little information and consisted mostly of formal language used in the parliamentary protocol.

	\begin{table}[H]
		\centering
        \small
			\begin{tabular}{lcccc}
    \toprule
    \textbf{Party} & \textbf{Precision} & \textbf{Recall} & \textbf{F1} & \textbf{Support} \\ \midrule
    		SPÖ & 0.61 & 0.64 & 0.62 & 24218 \\ 
    		ÖVP & 0.63 & 0.65 & 0.64 & 23428 \\ 
    		FPÖ & 0.55 & 0.66 & 0.60 & 16766 \\ 
    		GRÜNE & 0.67 & 0.53 & 0.59 & 12093 \\ 
    		LIF & 0.51 & 0.37 & 0.43 & 1869 \\ 
    		BZÖ & 0.40 & 0.34 & 0.36 & 4108 \\ 
    		NEOS & 0.55 & 0.36 & 0.44 & 1833 \\ 
    		STRONACH & 0.17 & 0.09 & 0.12 & 1298 \\ 
    		PILZ & 0.05 & 0.01 & 0.02 & 338 \\ \bottomrule
    		\end{tabular}
		\caption{Scores for speeches >30 tokens}
		\label{tab:ch5_scores_over30terms}
	\end{table}

The performance is better without those speeches as seen in Table ~\ref{tab:ch5_scores_over30terms}. 
The improvement is most pronounced for the smaller parties, which makes sense as the proportion of formal speeches is higher because they spend less time in parliament. 
Still the performance on speech level is not convincing.
We hypothesise that a single speech does not provide enough context to identify the party affiliation. Classifier failures could also be caused due to common topics between parties. Moreover, parties might have overlapping positions on individual topics but on others they might disagree strongly, making it difficult to tell them apart based on a single speech. To alleviate these problems, we next evaluate outputs aggregated at the speaker level.

\subsection{Evaluation at the level of the speaker}
\label{sub:res_table}
The speeches with >30 tokens were aggregated on speaker level with a simple mean, again choosing the maximum probability as the predicted party. 
As a result, 613 speakers were labelled fully correct.
From the switchers, 21 speakers were partially correct, meaning that one of the affiliated parties was predicted, but the predicted party was not the final party of that speaker. 
Out of those 21 wrongly labelled switching speakers, seven are completely wrong, not predicting either one of the parties the speaker belongs or belonged to. 

The performance immediately improves when looking at the score on this aggregated CM level, as shown in Table~\ref{tab:ch5_score_mp}. 
	
In Table ~\ref{tab:ch5_score_mp} the F1-score for all parties except PILZ and STRONACH are above 0.5, with the larger parties reaching between 0.85 and 0.9. 
\begin{table}[H]
    	\centering
        \small
    		\begin{tabular}{lccc}
    		    \toprule
    \textbf{Party} & \textbf{Precision} & \textbf{Recall} & \textbf{F1}  \\
    \midrule 
    		SPÖ & 0,89 & 0,89 & 0,89 \\ 
    		ÖVP & 0,85 & 0,95 & 0,90 \\ 
    		FPÖ & 0,86 & 0,90 & 0,88 \\ 
    		GRÜNE & 0,86 & 0,84 & 0,85 \\ 
    		LIF & 0,83 & 0,42 & 0,56 \\ 
    		BZÖ & 0,78 & 0,44 & 0,56 \\ 
    		NEOS & 0,90 & 0,64 & 0,75 \\ 
    		STRONACH & 1 & 0,17 & 0,29 \\ 
    		PILZ & 0 & 0 & 0 \\ \bottomrule
    		\end{tabular}
		\caption{Scores when aggregating the speech predictions the CM level}
		\label{tab:ch5_score_mp}
	\end{table}
After investigating the speeches on a CM level, it turned out that 74 CMs had less than ten speeches.
In order to use speakers that had more speaking time in Parliament, we excluded the 10\% of CMs with the fewest speeches. 
Using the labels provided by the data set, the model has an accuracy of 88.6\% for predicting the party affiliation of a speaker. This confirms that the model is of sufficient quality to be used in the analysis part of our approach (see Section~6.).

As mentioned before in preliminary experiments the authors trained a Logistic Regression Model using tf-idf weighted bag-of-words (BoW) features. Instead of splitting the data in five folds there was a model trained for each CM using leave-one-out cross-validation. Here too, the results improved when aggregated to the speaker level. However, as seen in Table~\ref{tab:ch5scoremplogreg}, the overall performance of that approach was worse than the BERT-based classifier. When manually examining the mislabelled speeches, it became clear that a BoW-based approach was unable to capture nuanced meaning differences and could lead to biases, as it relied more on specific topics being present rather than what was actually said. For example, speeches by a social-democratic CM mentioning the word "immigration" would be assigned to one of the right-wing parties, even if the speech itself was expressing positive views on immigration.

\begin{table}[H]
    	\centering
        \small
    		\begin{tabular}{lccc}
    		    \toprule
    \textbf{Party} & \textbf{Precision} & \textbf{Recall} & \textbf{F1}  \\
    \midrule 
    		SPÖ & 0,74 & 0,82 & 0,78 \\ 
    		ÖVP & 0,82 & 0,90 & 0,86 \\ 
    		FPÖ & 0,84 & 0,91 & 0,87 \\ 
    		GRÜNE & 0,77 & 0,68 & 0,72 \\ 
    		LIF & 1 & 0,36 & 0,53 \\ 
    		BZÖ & 1 & 0,11 & 0,19 \\ 
    		NEOS & 0,75 & 0,20 & 0,32 \\ 
    		STRONACH & 0 & 0 & 0 \\ 
    		PILZ & 0 & 0 & 0 \\ \bottomrule
    		\end{tabular}
		\caption{Scores when aggregating the speech predictions the CM level - Logistic Regression}
		\label{tab:ch5scoremplogreg}
	\end{table}

After investigating the speeches on a CM level, it turned out that 74 CMs had less than ten speeches.
In order to use speakers that had more speaking time in Parliament, we excluded the 10\% of CMs with the fewest speeches. 

Evaluated on the gold labels provided by the data set, the model has an accuracy of 88.6\% for predicting the party affiliation of a speaker. This confirms that the model is of sufficient quality to be used in the analysis part of our approach (please see Section~6.).

\subsection{Model failures}
Table ~\ref{tab:ch5_mp_cross_table} is contingency table showing the actual parties in the columns and the predicted parties in the rows. 
We can see that the predictions work quite well with a strong diagonal for the bigger parties. 

Only for the smaller parties STRONACH and PILZ  it gets more difficult for the model to classify. No CM was classified as PILZ and only two as STRONACH. This is in line with the smaller amount of training data as seen in Figure \ref{fig:ch3_2}.

\begin{table*}

\centering
\footnotesize
    \begin{tabular}{lccccccccc}
    	\toprule
        Party& SPÖ & ÖVP & FPÖ & GRÜNE & LIF & BZÖ & NEOS & STRONACH & PILZ\\ \midrule
        SPÖ & \textbf{185} & 19 & 0 & 2 & 0 & 0 & 1 & 0 & 0 \\ 
        ÖVP & 6 & \textbf{211} & 5 & 0 & 0 & 0 & 0 & 0 & 0\\ 
        FPÖ & 2 & 15 & \textbf{157} & 0 & 1 & 0 & 0 & 0 & 0\\ 
        GRÜNE & 6 & 1 & 0 & \textbf{37} & 0 & 0 & 0 & 0 & 0\\ 
        LIF & 3 & 0 & 2 & 2 & \textbf{5} & 0 & 0 & 0 & 0\\ 
        BZÖ & 0 & 1 & 8 & 0 & 0 & \textbf{7} & 0 & 0 & 0\\ 
        NEOS & 1 & 0 & 3 & 1 & 0 & 0 & \textbf{9} & 0 & 0\\ 
        STRONACH & 0 & 1 & 7 & 0 & 0 & 2 & 0 & \textbf{2} & 0\\ 
        PILZ & 5 & 0 & 0 & 1 & 0 & 0 & 0 & 0 & 0\\ \bottomrule
    	\end{tabular}
	\caption{Predicted party (columns) and actual party (rows)}
	\label{tab:ch5_mp_cross_table}
\end{table*}

We can also see that CMs are usually classified towards the "same" political direction. 
Meaning that a CM from a left party is rather misclassified to another left party and a CM from a right party is rather misclassified to another right party. 
Furthermore, the parties are often misclassified with their predecessor party (if they have one).
For example, a LIF CM or BZÖ CM might be misclassified as an FPÖ CM or a PILZ CM as GRÜNE. 
An extreme example is BZÖ where on CM is classified as ÖVP, seven as the correct label BZÖ and most CMs were classified as FPÖ with eight predicted CMs.

The final result of our classification efforts is a dataset of speeches with corresponding predicted probability distributions. In the next section we will show how the predictions along with model failure cases can be used to derive insights and guide qualitative analysis.

\section{Analysis and Discussion}

In this chapter we focus on the second step of our approach - analysis of the predicted probabilities to gain insights into the data.

\subsection{Visualising speakers}
\begin{figure}
	\centering
	\includegraphics[width=0.5\textwidth]{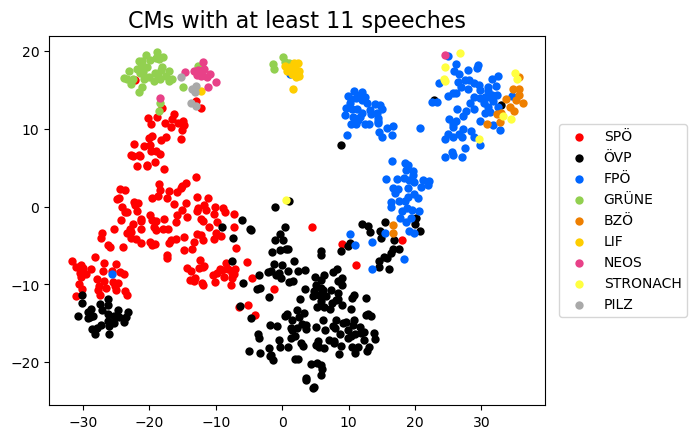}
	\caption{Relation of CMs to each other}
	\label{fig:ch5_tsne_mps}
\end{figure}

To visualize speaker relationships (Figure~\ref{fig:ch5_tsne_mps}), we project each speaker's party probability distribution to 2D using t-SNE \cite{van2008visualizing}. 
Distances between points indicate similarity; colors show actual party affiliations, with switched members assigned to their longer-held party.

We can see that the parties are clustered fairly well, further confirming the classifier captures meaningful patterns in the data. 
The orange dots on the bottom of the t-SNE visualisation are the BZÖ, which was created by former FPÖ members. 
The yellow dots represent LIF, they are socially liberal but economically right wing. 
The black dots that are separated by the red dots on top might be the "grand coalition" CMs of the 1990s. 
In that time the idea of ``Sozialpartnerschaft'' (\emph{``social partnership''}) was very strong. 
It meant that both conservatives and social democrats worked for the better good. 
GRÜNE is furthest from FPÖ and ÖVP which also corresponds to the expectation, even though there are some more economically liberal Greens that are clustered with LIF. 
The dark brown dots are the STRONACH party, they blend in with FPÖ, BZÖ and ÖVP which are the parties its CMs switched from. 

\subsection{Party affiliation}
\input{switchingaffiliationshort}

\begin{figure*}
    \centering
\includegraphics[width=0.9\textwidth]{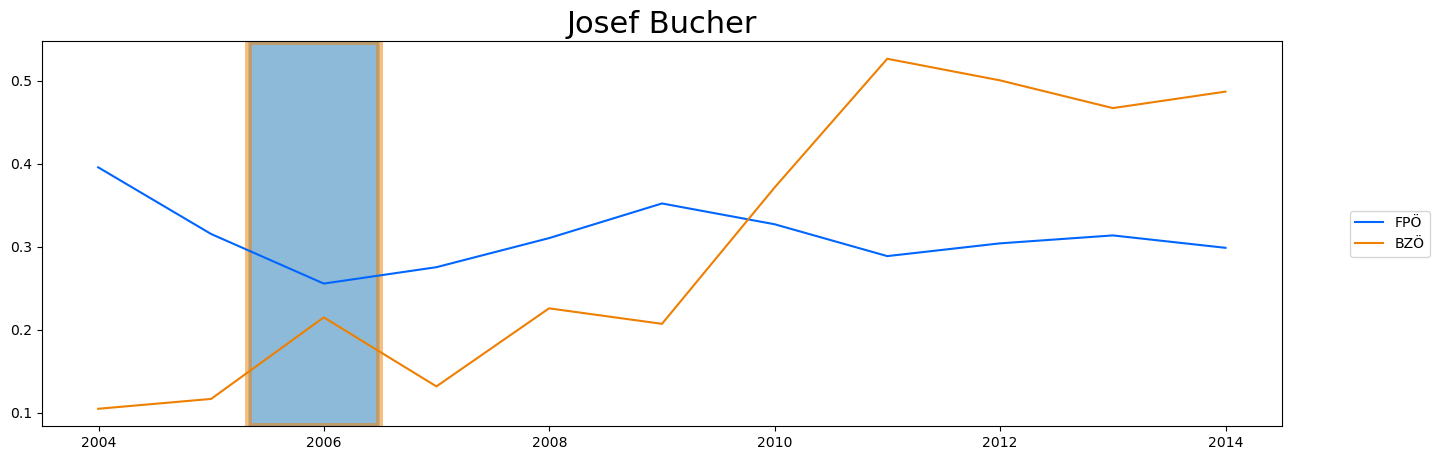}
	\caption[Probability of speeches being FPÖ and BZÖ]{Probability of speeches being FPÖ(blue) and BZÖ(orange) over time. The box shows the time of creation of the BZÖ and the slow rise in speeches being predicted as BZÖ over time.}
	\label{fig:ch6_josef_bucher}
\end{figure*}

\paragraph{Misclassifications} Another clear observation is that the misclassifications of the model are not "too far away" politically speaking, giving us further confidence that the model outputs are meaningful.
E.g., the left party GRÜNE is more likely to misclassify to SPÖ, the centre-left party, or SPÖ will most likely misclassify to GRÜNE or the centre-right party ÖVP. 
Only LIF and NEOS do not follow this pattern. They misclassify to both parties on the left and right, GRÜNE, SPÖ, FPÖ, ÖVP, because they are socially liberal but economically right. The models struggle to classify them clearly. 

Model mistakes can give important information about a CM's background which can be revealed by manual qualitative analysis. The SPÖ CM Ulrike Simma is briefly discussed as an example: 
Her misclassification is a reflection of her political background prior to running to become a CM for SPÖ in the 1999 elections. 
She had been a member of the green student party, as well as working for an environmentalist NGO (Global 2000). 
She ran for GRÜNE in the election in 1996 and only joined SPÖ for the election in 1999, when she got into Parliament as an CM for SPÖ. Please see the Appendix for a detailed description of all misclassified CMs.

\begin{figure*}
	\centering
	\includegraphics[width=0.95\textwidth]{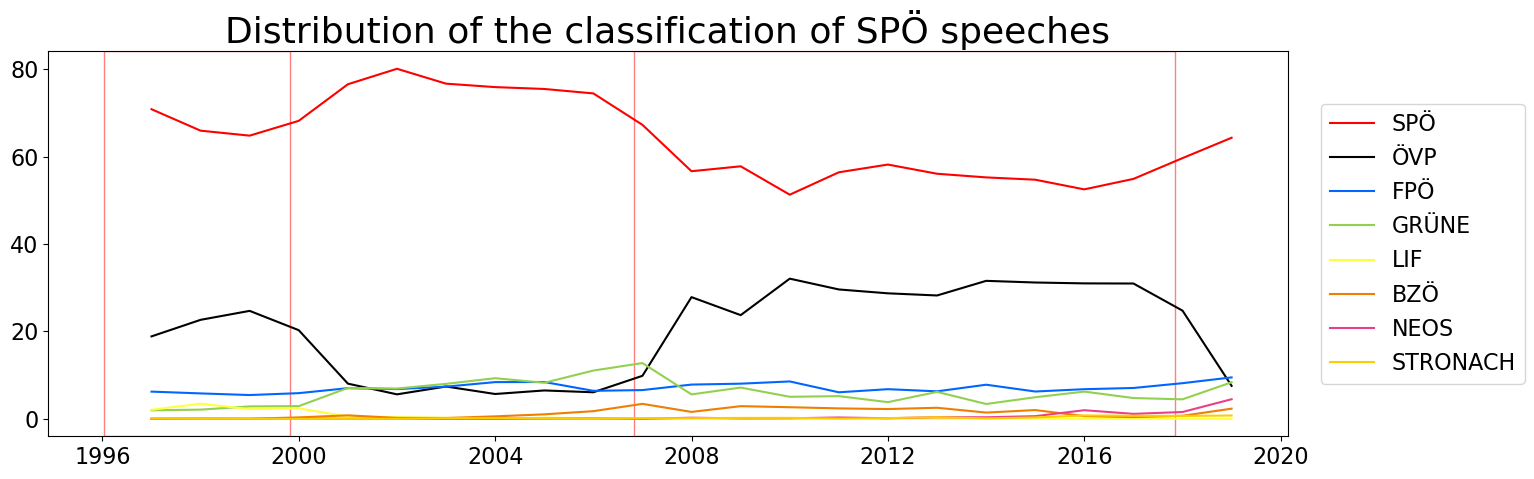}
	\caption{Proportion of SPÖ speeches being classified as ÖVP or FPÖ}
	\label{fig:ch6_misclassified_sp}
\end{figure*}
\begin{figure*}
	\centering
	\includegraphics[width=0.95\textwidth]{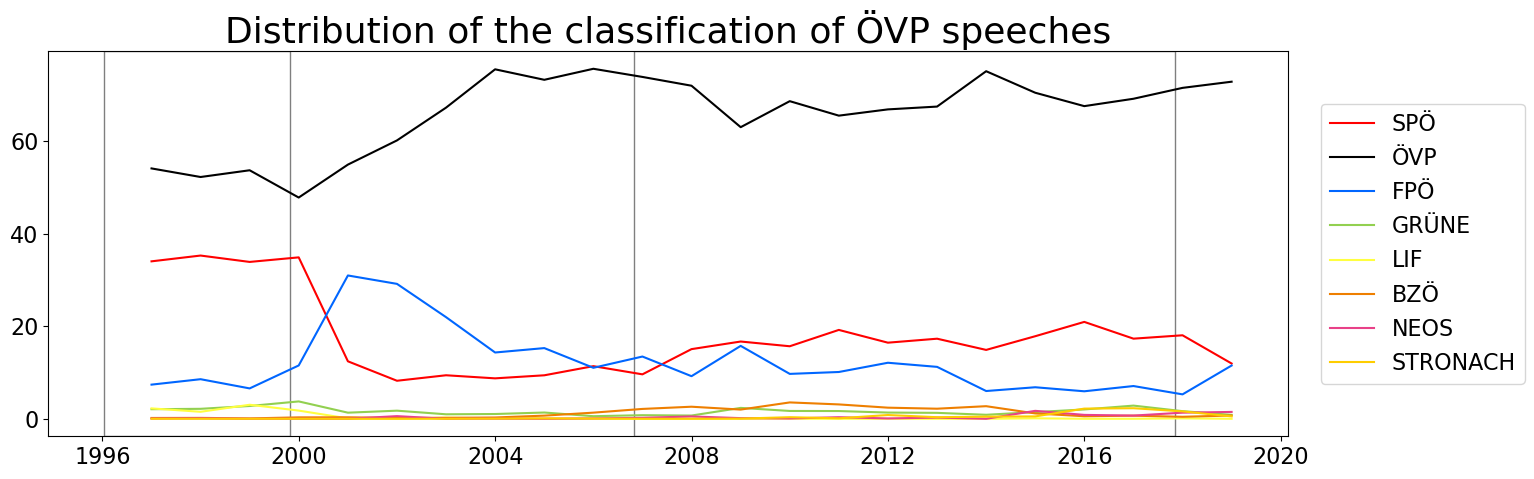}
	\caption{Proportion of ÖVP speeches being classified as ÖVP or FPÖ}
	\label{fig:ch6_misclassified_vp}
\end{figure*}
\begin{figure*}
	\centering
	\includegraphics[width=0.95\textwidth]{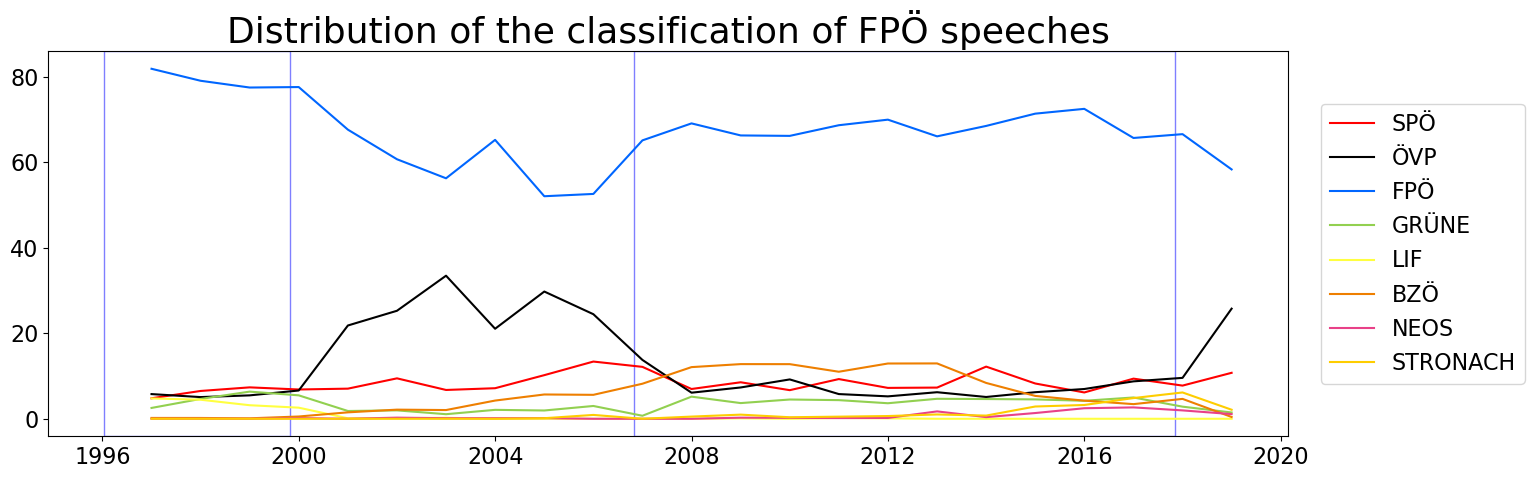}
	\caption{Proportion of FPÖ speeches being classified as ÖVP or FPÖ}
	\label{fig:ch6_misclassified_fp}
\end{figure*}
\begin{figure*}
	\centering
	\includegraphics[width=0.95\textwidth]{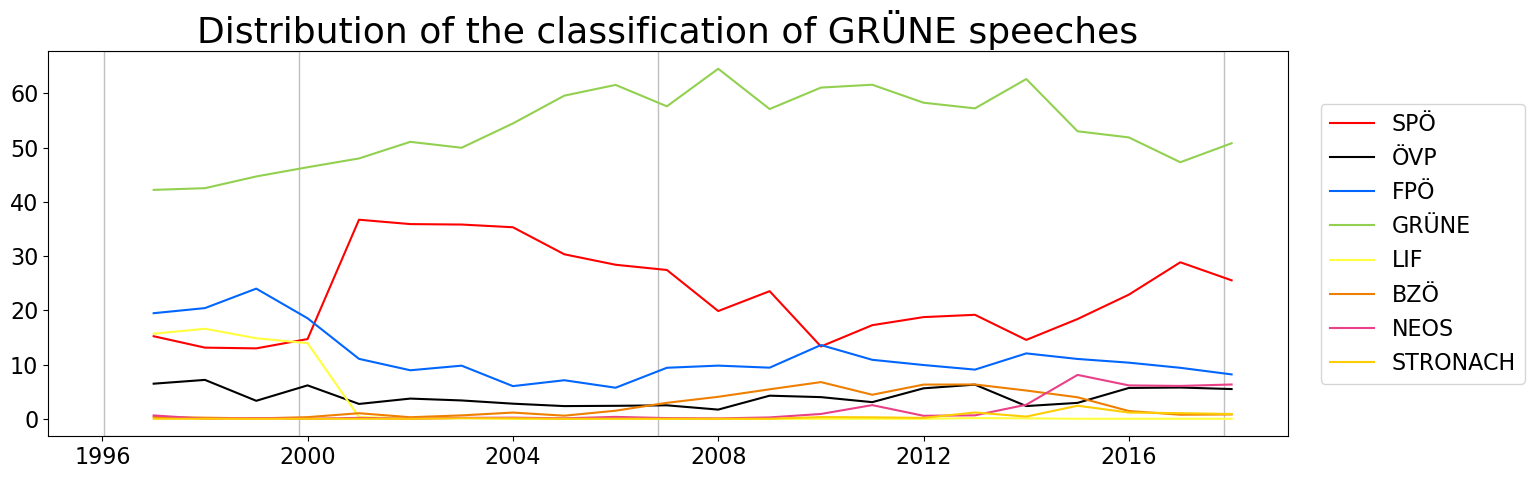}
	\caption{Proportion of GRÜNE speeches being classified as ÖVP or FPÖ}
	\label{fig:ch6_misclassified_gr}
\end{figure*}

\subsection{Government vs opposition}
We also analysed the model predictions/missclassifications over time. To this end we shortly describe the governments in the legislative periods covered in the data: firstly there was a grand coalition between SPÖ and ÖVP, then the first coalition between ÖVP and FPÖ (black-blue), followed by another three periods with a grand coalition between SPÖ and ÖVP and finally the last period in the data is the second black-blue coalition between ÖVP and FPÖ. ~\footnote{To remind the reader; SPÖ is the social democratic, centre-left party. ÖVP is the conservative, centre-right party and FPÖ is a right-wing party. GRÜNE were in government after the data set's time span and are a left-wing environmentalist party. } The timeline is shown below.

\vspace{1em}
\hspace*{-1.1em}
\begin{tikzpicture}[scale=0.75]
\draw[->] (-5.1,0) -- (4.5,0) node [below] {};
\foreach \x/\year/\text in {-4/1996/grand coalition,-1.5/1999/black-blue 1, 1/2006/grand coalition, 3.5/2017/black-blue 2}
\draw (\x, 0.1) -- (\x, -0.1) node [below] {\footnotesize {\shortstack{\year\\\text}}};
\end{tikzpicture}
 
Looking at the predictions of all speeches over time revealed that parties behave differently in opposition and government. 
When governing, they get closer to their coalition partner, whilst in opposition, they move away from the governing parties. 
This does not apply equally to all governing and oppositional parties at all times, but has a temporal aspect that affects some parties more than others:
Most misclassifications of SPÖ to ÖVP happened after the first coalition of FPÖ and ÖVP as shown in ~\ref{fig:ch6_misclassified_sp}.
That means that there was a difference in speech behaviour the second time SPÖ was in government with ÖVP compared to the first time they governed together in this data set. 
Most misclassifications from ÖVP to SPÖ happened before the first time FPÖ was in government with ÖVP.
After the first grand coalition ~\footnote{Grand coalition means the coalition between SPÖ and ÖVP, the traditionally biggest parties in Parliament. Today, this might not be a relevant or correct term, as FPÖ has grown to become just as large as the other two parties.} in the data set, ÖVP moved away from SPÖ and never fully returned to the proximity it had when it was in coalition the first time (see ~\ref{fig:ch6_misclassified_vp}. 
Also, the misclassifications of ÖVP to FPÖ happened within or after the first time working with FPÖ.

There is one more observation that may be of interest for further research in the field of political science.
ÖVP has never been in opposition in the data set and according to the models predictions, has moved more to the right.
In fact, looking at the models predicted probability distributions of each speech between 1996 and 2018, the Parliament as a whole has moved away from the left as there were no new left parties to take up the space in the political spectrum that ÖVP moved away from.
That means that there are fewer SPÖ labelled speeches overall in Parliament by time and no new parties that would give speeches that are classified towards the left, except for PILZ which replaced GRÜNE for one period.

Deducting from the previous finding that parties become more similar in their speeches when in coalition, analogously, when no longer in coalition parties should distance themselves again to the extent they got closer during governing times.
However, Because ÖVP was never in an oppositional position within the covered data it is difficult to verify if this is the case. 
It is still observable in Figure \ref{fig:ch6_misclassified_vp}, that the misclassified speeches towards SPÖ drastically reduced after being in a coalition with FPÖ. Similarly, looking at Figure~\ref{fig:ch6_misclassified_gr} we see how GRÜNE moves further away from the SPÖ during the times the SPÖ was in government and get closer to it when they are together in opposition.

\input{conclusionshort}

\input{limitationsshort}

\bibliography{custom}

\section{Appendix}
\label{sec:appendix}
\subsection{Misclassified CMs}
In this chapter I will discuss all the misclassified CMs where the model, in predicting their speeches, failed to correctly assign the party that they belong to. 
The aim is to find patterns in the misclassifications and use them for interpretation. 

\subsubsection{SPÖ}
Table~\ref{tab:ch6_spö_misclassified} shows that by aggregating the speech predictions on a personal level 22 SPÖ CMs were misclassified as belonging to a different party. The final column in the table shows the predicted party for each of them based on the highest party affiliation score from the model. 19 were predicted to be ÖVP CMs, two GRÜNE and one NEOS.

	\begin{table*}
	\small
    \centering
    \begin{tabular}{|p{3cm}|l|l|l|l|l|l|l|l|l|l|}
    \hline
        Speaker & SPÖ & ÖVP & FPÖ & GRÜNE & LIF & BZÖ & NEOS & TS & PILZ & predicted \\ \hline
        Alexander Zach & 0.26 & 0.07 & 0.05 & 0.23 & 0.04 & 0.01& 0.26 & 0.01 & 0.06 & NEOS  \\ \hline
        Heinz Fischer & 0.22 & 0.14 & 0.15 & 0.45 & 0.04& 0.00 & 0.00 & 0.00 & 0.00 & GRÜNE  \\ \hline
        Ulrike Sima & 0.29 & 0.04 & 0.16 & 0.47 & 0.00 & 0.03 & 0.00 & 0.00 & 0.00 & GRÜNE \\ \hline
		Alois Stöger & 0.37 & 0.53 & 0.07 & 0.01 & 0.00 & 0.01 & 0.00 & 0.00 & 0.00 & ÖVP \\ \hline
        Andrea \mbox{Gessl-Ranftl} & 0.42 & 0.44 & 0.06 & 0.03 & 0.00 & 0.03 & 0.00 & 0.01 & 0.00& ÖVP  \\ \hline
        Christian Hursky & 0.32 & 0.46 & 0.14 & 0.04 & 0.00 & 0.02 & 0.01 & 0.00 & 0.00 & ÖVP \\ \hline
        Claudia Schmied & 0.36 & 0.59 & 0.02 & 0.01 & 0.00 & 0.02 & 0.01 & 0.00 & 0.00 & ÖVP \\ \hline
        Gerald Klug & 0.39 & 0.46 & 0.13 & 0.01 & 0.00 & 0.00 & 0.00 & 0.00 & 0.00 & ÖVP \\ \hline
        Gerhard Köfer & 0.27 & 0.38 & 0.17 & 0.00 & 0.00 & 0.15 & 0.00 & 0.02 & 0.00 & ÖVP \\ \hline
        Harald Troch & 0.34 & 0.35 & 0.15 & 0.08 & 0.00 & 0.04 & 0.02 & 0.00 & 0.01 & ÖVP \\ \hline
        Johann Hechtl & 0.35 & 0.55 & 0.06 & 0.00 & 0.00 & 0.02 & 0.00 & 0.00 & 0.00 & ÖVP  \\ \hline
        Josef Auer & 0.28 & 0.29 & 0.24 & 0.04 & 0.00 & 0.05 & 0.05 & 0.04 & 0.00 & ÖVP \\ \hline
        Jürgen Schabhüttl & 0.28 & 0.57 & 0.1 & 0.01 & 0.00 & 0.05 & 0.00 & 0.00 & 0.00 & ÖVP \\ \hline
        Maria Berger & 0.33 & 0.5 & 0.03 & 0.04 & 0.00 & 0.09 & 0.00 & 0.00 & 0.00 & ÖVP  \\ \hline
        Michael Ehmann & 0.39 & 0.42 & 0.11 & 0.03 & 0.00 & 0.01 & 0.02 & 0.01 & 0.00 & ÖVP \\ \hline
        Michael Schickhofer & 0.37 & 0.49 & 0.08 & 0.03 & 0.00 & 0.02 & 0.01 & 0.01 & 0.00 & ÖVP \\ \hline
        Norbert Darabos & 0.27 & 0.37 & 0.28 & 0.04 & 0.00 & 0.03 & 0.01 & 0.00 & 0.00 & ÖVP  \\ \hline
        Peter Doskozil & 0.14 & 0.46 & 0.37 & 0.01 & 0.00 & 0.01 & 0.00 & 0.00 & 0.00 & ÖVP \\ \hline
        Philip Kucher & 0.35 & 0.43 & 0.06 & 0.06 & 0.00 & 0.03 & 0.03 & 0.02 & 0.02 & ÖVP \\ \hline
        Rudolf Hundstorfer & 0.38 & 0.43 & 0.1 & 0.05 & 0.00 & 0.02 & 0.01 & 0.00 & 0.00 & ÖVP  \\ \hline
        Rudolf Plessl & 0.33 & 0.43 & 0.16 & 0.03 & 0.00 & 0.02 & 0.01 & 0.02 & 0.01 & ÖVP  \\ \hline
        Sonja \mbox{Steßl-Mühlbacher} & 0.34 & 0.44 & 0.12 & 0.05 & 0.00 & 0.03 & 0.01 & 0.00 & 0.01 & ÖVP \\ \hline
	\end{tabular}
	\caption{Party affiliation score of misclassified SPÖ members}
	\label{tab:ch6_spö_misclassified}
	\end{table*}

	\subsubsubsection{SPÖ → NEOS}
	Alexander Zach was actually a member of LIF. For the 2006 elections LIF cooperated with SPÖ with the aim to prevent a right wing majority between ÖVP and FPÖ. 
	For practical reasons he was part of SPÖ Parliamentary Group, but politically he was completely independent. 
	LIF merged with NEOS in 2014 so his prediction as NEOS member is not too far away. 
	He was sent to the ORF Board of Trustees by NEOS.
	
	\subsubsubsection{SPÖ → GRÜNE}

	Two SPÖ CMs were misclassified as GRÜNE, Ulrike Sima and Heinz Fischer.
	Ulrike Sima's misclassification is a reflection of her political background prior to running to become an CM for SPÖ in the  1999 elections. 	
	She had been a member of the green student party, as well as working for an environmentalist NGO (Global 2000).
	She ran for GRÜNE in the election in 1996 and only joined SPÖ for the election in 1999, when she got into Parliament as an CM for SPÖ. 
	Heinz Fischer has only two speeches with more than 30 words in the data set. 
	Most of his speeches were held in the position of the President of the Parliament or even the President of Austria. 
	Those speeches were labelled one as GRÜNE and one as ÖVP. 
	One speech is more administrative.
	In that speech he responds to an accusation of his organisational role as a President of the Parliament. 
	Two speeches might be too few to say what party a politician belongs to.

	\subsubsubsection{SPÖ → ÖVP}

	The 19 SPÖ CMs who were misclassified as ÖVP can be separated into two groups.
	The first group includes those CMs who were appointed Ministers by the SPÖ when in coalition with ÖVP.
	This is by far the biggest group of CMs that has been misclassified with 19 CMs.
	The second group consists of CMs that joined after the first FPÖ-ÖVP government, when SPÖ was in coalition with ÖVP again. 
	The Ministers that were in an SPÖ-ÖVP government before the first ÖVP-FPÖ coalition are Alois Stöger, Claudia Schmied, Maria-Margarethe Berger, Rudolf Hundstorfer and Norbert Darabos.
	Because those Ministers are in a coalition with the ÖVP, they might use similar language to argue for policies and agendas that are set by the SPÖ-ÖVP government. 
	The need to argue for a common governmental policy could be a possible explanation for the misclassification of the speeches in this group of ministers.
	As an example for those Ministers I will discuss Claudia Schmied in more detail.
	Claudia Schmied was a Minister under SPÖ-ÖVP coalitions from January 2007 to December 2013. 
	Most of her speeches were labelled ÖVP, but there are many SPÖ labelled speeches too.
	She often deals with topics around her Ministry of Education.
	It is remarkable that speeches with roughly the same topic and length were labelled differently.
	One was labelled strongly as SPÖ and one was labelled strongly as ÖVP. 
	The speech about all-day child care where she emphasises the freedom of choice for parents was classified as ÖVP. 
	A speech talking about the exact same topic with the argument that all-day child care helps to fight social inequalities was labelled SPÖ. 
	This is an interesting example of two speeches speaking from the position of the government, but using different ideological view points to argue for the political agenda of all-day child care.

	The Ministers that were in an SPÖ-ÖVP government after the first ÖVP-FPÖ government are Norbert Darabos, Hans Peter Doskozil, Rudolf Hundstorfer, Sonja Steßl and Gerald Klug.
	The fact that they were Ministers in a government coalition with ÖVP could hint towards SPÖ governing politicians adapting their speech behaviour towards their coalition partner ÖVP. 

	\begin{figure*}
	\centering
	\includegraphics[width=1\textwidth]{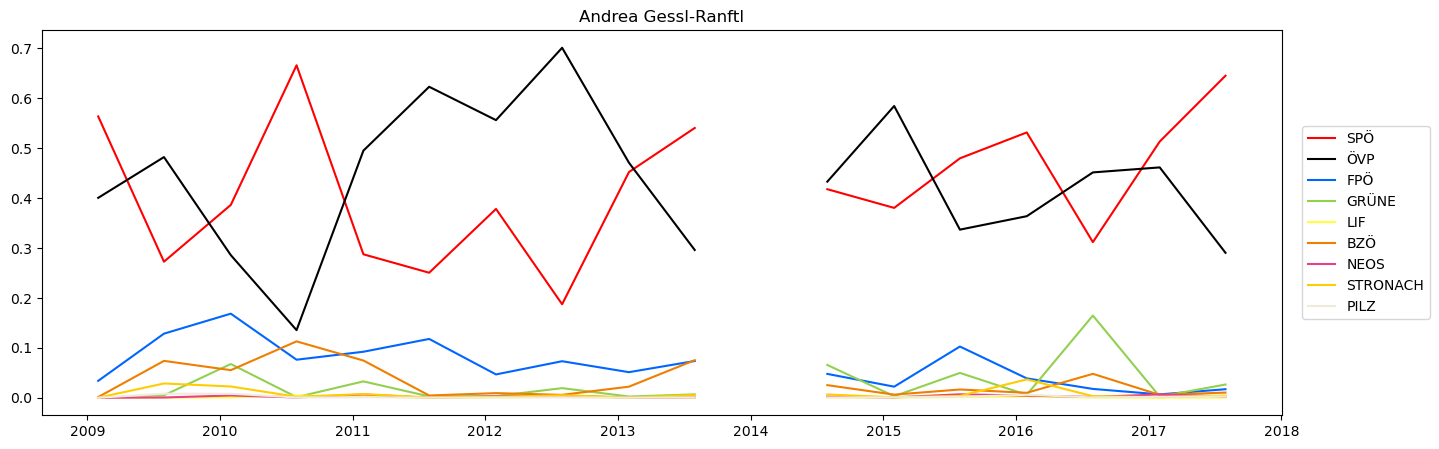}
	\caption{Probability distribution of speeches}
	\label{fig:ch6_andrea_gessl_ranftl}
	\end{figure*}

	There are similar dynamics amongst CMs that got into Parliament during a "grand coalition":
	Andrea Gessl-Ranftl, Christian Hursky, Johann Hechtl, Harald Troch, Josef Auer, Jürgen Schabhüttel, Michael Schickhofer, Michael Ehmann, Philip Kucher and Rudolf Plessel all got into Parliament in a coalition period with the ÖVP.

	Andrea Gnessl-Ranftl was in Parliament for two legislative periods, both were "grand coalitions". 
	As seen in Figure ~\ref{fig:ch6_andrea_gessl_ranftl} her speech behaviour moved a bit more towards SPÖ when she was an CM for the second time. 
	In the speeches that were strongly labelled ÖVP, she does comment on things often in a defensive manner.
	Either she explains what the government is already doing regarding a topic, or she is mentioning and explaining the complexity of certain demands. 

	Some of the SPÖ CMs misclassified as ÖVP were in Parliament for more than just the periods that SPÖ was in government.
	Harald Troch was in Parliament from October 2013 until October 2024.
	He experienced being an CM both in government and in opposition. 
	His aggregated score for ÖVP and SPÖ is between 34-35\%. 
	Philip Kucher came into Parliament in October 2013 and is still there. 
	With those CMs the speech behaviour of the coalition times might have influenced their oppositional speech behaviour as well.~\footnote{It would be interesting to gather more data about how influential the first period in Parliament is for CMs and if the first period of an CM in Parliament shapes their speech behaviour.}
	
	\subsubsubsection{Gehard Köfer}
	He left SPÖ and became independent for the last nine months of his time in Parliament.
	In that time he joined the newly founded party Team Stronach. 
	He switched to the federal government with his new party, before STRONACH became its own Parliamentary Group, which is why he was independent as an CM even though he already was a party member of STRONACH.
	That is why his speeches are labelled independent. 
	His speeches after he left SPÖ are either FPÖ or ÖVP, which would fit the misclassification as ÖVP.
	After some scandals and conflicts with STRONACH, he founded his own regional party in Carinthia. 

	\subsubsection{ÖVP}

	ÖVP has eleven misclassified CMs as seen in Table ~\ref{tab:ch6_övp_misclassified_mps}.
	There are no oppositional ÖVP speeches as they were always in government in the timespan of the data set. 
	Six CMs are misclassified as SPÖ and five are wrongly labelled as FPÖ.

	\begin{table*}
    \centering
    \small
    \begin{tabular}{|p{3cm}|l|l|l|l|l|l|l|l|l|l|}
    \hline
        Speaker & SPÖ & ÖVP & FPÖ & GRÜNE & LIF & BZÖ & NEOS & TS & PILZ & predicted\\
		Franz Steindl & 0.38 & 0.12 & 0.04 & 0.03 & 0.00 & 0.00 & 0.00 & 0.00 & 0.00 & SPÖ\\ \hline
        Hannes Farnleitner & 0.33 & 0.03 & 0.01 & 0.05 & 0.00 & 0.00 & 0.00 & 0.00 & 0.00 & SPÖ \\ \hline
        Joseph Huainigg & 0.3 & 0.1 & 0.14 & 0.00 & 0.07 & 0.01 & 0.03 & 0.01 & 0.00 & SPÖ \\ \hline
        Juliane Bogner-Strauß & 0.4 & 0.06 & 0.03 & 0.00 & 0.02 & 0.02 & 0.01 & 0.01 & 0.00 & SPÖ \\ \hline
        Kira Grünberg & 0.34 & 0.12 & 0.06 & 0.00 & 0.01 & 0.01 & 0.01 & 0.02 & 0.00 & SPÖ \\ \hline
        Werner Fasslabend & 0.34 & 0.15 & 0.02 & 0.02 & 0.01 & 0.00 & 0.00 & 0.00 & 0.00 & SPÖ \\ \hline
		Bernhard Vock & 0.15 & 0.4 & 0.05 & 0.00 & 0.23 & 0.02 & 0.04 & 0.01 & 2 & FPÖ \\ \hline
        Ernst Fink & 0.32 & 0.37 & 0.01 & 0.01 & 0.01 & 0.00 & 0.00 & 0.00 & 2 & FPÖ \\ \hline
        Gerhart Bruckmann & 0.41 & 0.46 & 0.01 & 0.00 & 0.01 & 0.00 & 0.00 & 0.00 & 2 & FPÖ \\ \hline
        Helmut Kukacka & 0.32 & 0.36 & 0.04 & 0.01 & 0.03 & 0.00 & 0.00 & 0.00 & 2 & FPÖ \\ \hline
        Peter Fichtenbauer & 0.12 & 0.47 & 0.12 & 0 & 0.07 & 0.01 & 0.01 & 0.00 & 2 & FPÖ \\ \hline
	\end{tabular}
	\caption{Party affiliation score of misclassified ÖVP members}
	\label{tab:ch6_övp_misclassified_mps}
	\end{table*}
	\subsubsubsection{ÖVP → SPÖ}
	Three of the six ÖVP CMs misclassified as SPÖ CMs were in Parliament in the 1990s while ÖVP was in coalition with SPÖ.
	These CMs are Franz Steindl, Hannes Farnleitner and to an extent Werner Fasslabend.
	Franz Steindl was a member of the ÖAAB which is the workers association of ÖVP. 
	He was also active in the AK, which is the Workers Chamber. 
	His topics on workers rights might be the reason many of his speeches were classified as SPÖ, even though his second-strongest party is ÖVP.
	Hannes Farnleitner was Minister for Economy from 1996 to 2000 whilst SPÖ was in government with ÖVP.
	Werner Fasslabend was Minister for Defence from 1990 to 2000 and Third Parliamentary President from 2000 to 2002.
	As for the three other ÖVP CMs misclassified as SPÖ, Juliane Bogner-Strauß was Minister for Women from 2017 to 2018; Joseph Huainigg was the speaker for disabled people within the ÖVP Parliamentary Group; and Kira Grünberg is an CM who speaks about inclusive sport/society from her own experience as a professional sports person with disabilities.
	The last two misclassifications might be due to both speakers emphasising the importance of inclusive sports as it may be that SPÖ talks more about including groups with disadvantages.

	\subsubsubsection{ÖVP → FPÖ}

	As mentioned in the chapter on data quality, two CMs were labelled as ÖVP in the Parliamentary protocols even though they were FPÖ members: 
	Bernhard Vock and Peter Fichtenbauer. 

	The other three ÖVP CMs that are labelled as FPÖ by the model are Ernst Fink, Helmut Kukacka and Gerhart Bruckmann. 
	In Ernst Fink's speeches, he campaigns against the ÖBB (Austrian Railway) and the railway workers rights and their pension system. 
	Helmut Kukacka criticises the ÖBB in his speeches too. 
	He was head of the Mittelschul-Kartell-Verband (Christian fraternity). 
	As an oppositional party, FPÖ criticises governmental structures with regards to efficiency.
	Gerhart Bruckmann was the first pensioner CM. 
	He was a long working researcher at IHS before becoming an CM.  
	The FPÖ often push for privatisation and austerity politics. 

	\subsubsection{FPÖ}

	As Table ~\ref{tab:ch6_fpö_misclassfied_mps} shows 18 FPÖ CMs were misclassified, 15 as ÖVP, two as SPÖ and one as LIF.
	There are many CMs that actually switched party to the BZÖ, STRONACH or ÖVP. 

	\begin{table*}
    \centering
    \small
    \begin{tabular}{|p{3cm}|l|l|l|l|l|l|l|l|l|l|}
    \hline
        Speaker & SPÖ & ÖVP & FPÖ & GRÜNE & LIF & BZÖ & NEOS & TS & PILZ & predicted\\
		Franz Löschnak & 0.61 & 0.14 & 0.12 & 0.05 & 0.07 & 0.00 & 0.00 & 0.00 & 0.00 & SPÖ\\ \hline
        Hans Müller & 0.39 & 0.24 & 0.36 & 0.00 & 0.00 & 0.00 & 0.00 & 0.00 & 0.00 & SPÖ\\ \hline
        Alois Kainz & 0.3 & 0.34 & 0.33 & 0.01 & 0.00 & 0.01 & 0.01 & 0.01 & 0.00 & ÖVP\\ \hline
        Christian Pewny & 0.16 & 0.43 & 0.38 & 0.02 & 0.00 & 0.01 & 0.00 & 0.00 & 0.00 & ÖVP\\ \hline
        Christian Ragger & 0.26 & 0.44 & 0.18 & 0.00 & 0.00 & 0.04 & 0.06 & 0.00 & 0.00 & ÖVP\\ \hline
        Christian Ries & 0.14 & 0.45 & 0.32 & 0.00 & 0.00 & 0.06 & 0.00 & 0.01 & 0.01& ÖVP \\ \hline
        Christian Schandor & 0.24 & 0.44 & 0.29 & 0.01 & 0.00 & 0.01 & 0.00 & 0.00 & 0.00 & ÖVP\\ \hline
        Dieter Böhmdorfer & 0.15 & 0.34 & 0.25 & 0.03 & 0.01 & 0.22 & 0.00 & 0.00 & 0.00 & ÖVP\\ \hline
        Elisabeth Sickl & 0.1 & 0.46 & 0.42 & 0.01 & 0.00 & 0.01 & 0.00 & 0.00 & 0.00 & ÖVP\\ \hline
        Elmar Lichtenegger & 0.2 & 0.37 & 0.34 & 0.03 & 0.00 & 0.04 & 0.00 & 0.02 & 0.00& ÖVP \\ \hline
        Gerhard Hetzl & 0.18 & 0.41 & 0.39 & 0.01 & 0.00 & 0.01 & 0.00 & 0.00 & 0.00 & ÖVP\\ \hline
        Heinz Grasser & 0.16 & 0.59 & 0.22 & 0.00 & 0.00 & 0.02 & 0.00 & 0.00 & 0.00 & ÖVP\\ \hline
        Herbert \mbox{L. Graf} & 0.23 & 0.43 & 0.3 & 0.03 & 0.00 & 0.01 & 0.00 & 0.00 & 0.00 & ÖVP\\ \hline
        Marialuise Mittermüller & 0.13 & 0.46 & 0.28 & 0.00 & 0.00 & 0.12 & 0.00 & 0.00 & 0.00 & ÖVP\\ \hline
        Monika Forstinger & 0.32 & 0.38 & 0.25 & 0.00 & 0.00 & 0.05 & 0.00 & 0.00 & 0.00 & ÖVP\\ \hline
        Petra Wagner & 0.18 & 0.37 & 0.3 & 0.01 & 0.00 & 0.11 & 0.00 & 0.02 & 0.00 & ÖVP\\ \hline
        Roland Zellot & 0.12 & 0.43 & 0.41 & 0.02 & 0.00 & 0.02 & 0.00 & 0.00 & 0.00 & ÖVP \\ \hline
		Willi Brauneder & 0.19 & 0.13 & 0.25& 0.16 & 0.26 & 0.00 & 0.00 & 0.00 & 0.00 & LIF \\ \hline
	\end{tabular}
	\caption{Party affiliation score of misclassified FPÖ members}
	\label{tab:ch6_fpö_misclassfied_mps}
	\end{table*}

	\subsubsubsection{FPÖ → SPÖ}
	Franz Löschnak was actually an SPÖ CM and was wrongly labelled in the data. 
	Hans Müller was an CM for FPÖ from 1999 to 2002. 
	He speaks about economical and fiscal topics and often criticises the government. 
	\subsubsubsection{FPÖ → ÖVP}
	This is the biggest group of misclassified FPÖ CMs with 15 labelled as ÖVP. 
	Some of them were Ministers or CMs during a coalition between FPÖ and ÖVP~\footnote{They were Ministers in either one or more of the three legislative periods 1999 to 2002, 2002 to 2005 or 2017 to 2019.}. 

	Beate Hartinger-Klein occurs in the data set twice; first as Beate Hartinger and later as Beate Hartinger-Klein. 
	She was labelled correctly as FPÖ in her first period in Parliament from 1999 to 2002.
	In her second period she was wrongly labelled as ÖVP. 
	This is an example of a Minister being misclassified as the party they are in government with, as ÖVP and FPÖ were coalition partners at that time. 
	Dieter Böhmdorfer was Minister of Justice from 2000 to 2004 which was also during a time of coalition with ÖVP. 
	Elisabeth Sickl was Minister of Social Security and Generations and later Minister of Work, Health and Social Agenda, again when ÖVP was FPÖ's coalition partner. 
	Monika Forstinger was Minister for Transport for FPÖ from 2000 to 2002 in that "black-blue" coalition. 
	Roland Zellot was an CM for FPÖ from 1999 to 2005, covering a large part of FPÖs time in government.
	His predictions for FPÖ are very strong but he is still wrongly labelled as ÖVP.
	Petra Wanger was in Parliament from 2017 to 2019, again during a time when FPÖ was in government with ÖVP. 
	Alois Kainz was an CM from 2017 to 2019 and also is connected to the military. 

	There is a group of FPÖ CMs whose misclassification as ÖVP may relate to their speeches on the economy.
	This group includes Christian Pewny, Gerhard Hetzl and Herbert L. Graf.
	Christian Pewny who was an CM from 2017 to 2019 and a Member of the WKO~\footnote{The WKO, Wirtschaftskammer Österreich, is an institutionalised representation of industry, business and employers. Austria has a so called "chamber system", which has a Workers chamber "Arbeiterkammer (AK)" as well as chambers for doctors, farmers and other parts of society.}, has only eight speeches, four labelled FPÖ and only three ÖVP.
	His ÖVP speeches have a very high score for ÖVP.
	One of the speeches is about economical espionage and how to protect Austrian economy and innovation.
	Gerhard Hetzl an CM for FPÖ from 2000 to 2002 held 18 speeches, some of which were strongly labelled as ÖVP.
	Herbert L. Graf CM for FPÖ 1999 to 2002 was also a board member of the WKO. 
	With these three CMs it could be that their agenda to speak about the economy and business influenced their speech behaviour, particularly because the WKO is ideologically close to ÖVP.
	Christian Ragger an CM from 2017 to 2018 has only eleven speeches, again with two speeches that were strongly labelled as ÖVP.
	Christian Schandor an CM from 2017 to 2019 has only eleven speeches, and some were strongly labelled as ÖVP when he was talking about the economy.

	A third group of FPÖ CMs misclassified as ÖVP speakers are the ones that switched to the BZÖ. 
	Elmar Lichtenegger was an CM for FPÖ from 2003 to 2006 and later switched to the BZÖ in April 2006.
	Marialuise Mittermüller was an CM for FPÖ from 2005 to 2006.
	She has 29 speeches and also joined BZÖ in April 2006. 
	Heinz Grasser left FPÖ in 2003, but as an independent CM remained Minister of Finance nominated by the ÖVP, up until 2007.
	He distanced himself from both FPÖ and the BZÖ, but the parties also distanced themselves from him.

	\subsubsubsection{FPÖ → LIF}
	Willi Brauneder, who was wrongly classified as LIF, was Third President of Parliament.
	Ten of his 26 speeches were labelled as LIF, a few others as SPÖ and ÖVP and only eight as FPÖ. 
	He was Dean of the Law Faculty at the University of Vienna.

	\subsubsection{GRÜNE}
	\begin{table*}
    \small
    \centering
    \begin{tabular}{|p{3cm}|l|l|l|l|l|l|l|l|l|l|}
    \hline
        Speaker & SPÖ & ÖVP & FPÖ & GRÜNE & LIF & BZÖ & NEOS & TS & PILZ & predicted\\ \hline
        Berivan Aslan & 0.32 & 0.12 & 0.09 & 0.23 & 0.00 & 0.04 & 0.12 & 0.03 & 0.05 & SPÖ \\ \hline
        Bettina Hradecsni & 0.33 & 0.15 & 0.18 & 0.26 & 0.00 & 0.04 & 0.02 & 0.02 & 0.01 & SPÖ \\ \hline
        Heidemarie \mbox{Rest-Hinterseer} & 0.37 & 0.08 & 0.13 & 0.36 & 0.00 & 0.03 & 0.03 & 0.01 & 0.00 & SPÖ \\ \hline
        Johannes Voggenhuber & 0.49 & 0.09 & 0.03 & 0.22 & 0.17 & 0.00 & 0.00 & 0.00 & 0.00 & SPÖ \\ \hline
        Kurt Grünewald & 0.28 & 0.14 & 0.2 & 0.25 & 0.01 & 0.06 & 0.02 & 0.03 & 0.01 & SPÖ \\ \hline
        Wolfgang Pirklhuber & 0.4 & 0.08 & 0.1 & 0.33 & 0.01 & 0.05 & 0.01 & 0.01 & 0.01 & SPÖ \\ \hline
        Barbara Neuroth & 0.24 & 0.35 & 0.2 & 0.02 & 0.00 & 0.03 & 0.00 & 0.15 & 0.00 & ÖVP \\ \hline
    \end{tabular}
	\caption{Party affiliation score of misclassified GRÜNE members}
	\label{tab:ch6_gr_misclassified_mps}
	\end{table*}

	\subsubsubsection{GRÜNE → SPÖ}
	Table ~\ref{tab:ch6_gr_misclassified_mps} shows that six of the seven misclassified GRÜNE CMs were labelled as SPÖ. 
	The six CMs are Berivan Aslan, Bettina Hradecsni, Heidemarie Rest-Hinterseer, Johannes Voggenhuber (who ran for PILZ in the EU-elections 2019), Kurt Grünewald and Wolfgang Priklhuber (who talks about agriculture and food). 
	\subsubsubsection{GRÜNE → ÖVP}
	Barbara Neuroth, the only GRÜNE CM misclassified as ÖVP, was in Parliament for only five months with only two speeches. 
	That might be too little to actually predict a party affiliation. 
	\subsubsection{LIF}
	
	Table ~\ref{tab:ch6_lif_misclassified_mps} shows that LIF had seven misclassified CMs, three of whom were labelled as SPÖ, two as FPÖ and two as GRÜNE. 
	\begin{table*}
    \small
    \centering
    \begin{tabular}{|p{3cm}|l|l|l|l|l|l|l|l|l|l|}
    \hline
        Speaker & SPÖ & ÖVP & FPÖ & GRÜNE & LIF & BZÖ & NEOS & TS & PILZ & predicted \\ \hline
        Friedhelm Frischenschlager & 0.26 & 0.11 & 0.24 & 0.16 & 0.23 & 0 & 0 & 0 & 0 & SPÖ \\ \hline
        Helmut Moser & 0.36 & 0.16 & 0.19 & 0.07 & 0.2 & 0.01 & 0 & 0 & 0 & SPÖ \\ \hline
        Helmut Peter & 0.19 & 0.17 & 0.29 & 0.12 & 0.22 & 0 & 0 & 0 & 0 & FPÖ \\ \hline
        Maria Schaffenrath & 0.19 & 0.09 & 0.24 & 0.25 & 0.22 & 0.01 & 0 & 0 & 0 & GRÜNE \\ \hline
        Peter Haselsteiner & 0.15 & 0.11 & 0.31 & 0.14 & 0.3 & 0 & 0 & 0 & 0 & FPÖ \\ \hline
        Rosa Mlinar & 0.29 & 0.26 & 0.05 & 0.03 & 0.01 & 0.02 & 0.28 & 0.03 & 0.04 & SPÖ \\ \hline
        Thomas Barmüller & 0.19 & 0.07 & 0.19 & 0.29 & 0.26 & 0 & 0 & 0 & 0 & GRÜNE \\ \hline
    \end{tabular}
	\caption{Party affiliation score of misclassified LIF members}
	\label{tab:ch6_lif_misclassified_mps}
	\end{table*}
	\subsubsubsection{LIF → SPÖ}
	Of the three who were misclassified as SPÖ, Friedhelm Frischenschlager was an CM for FPÖ from 1977 to 1993 and Minister for Defence during the SPÖ minority government. 
	He became an CM for LIF later from 1993 to 1996.

	Hans Moser was also an CM for FPÖ from 1989 to 1993 and then switched to  LIF from 1993 to 1999.
	He is in the Austrian Military. 

	Rosa Mlinar was an CM for NEOS/LIF from 2013 to 2014 and a Member of the EU Parliament for NEOS from 2014 to 2019.
	Her speeches are either labelled NEOS or SPÖ. 

	\subsubsubsection{LIF → FPÖ}

	Of the two CMs that were labelled as FPÖ, Helmut Peter was actually an CM for FPÖ from 1990 to 1993 before joining LIF in 1994 where he stayed until 1999.
	Peter Haselsteiner was an CM for LIF from 1994 to 1998.
	He was a businessman who got involved in politics.

	\subsubsubsection{LIF → GRÜNE}
	Thomas Barmüller was an CM for FPÖ from 1990 to 1993 and for LIF from 1993 to 1999.
	Maria Schaffenrath was an CM for LIF from 1994 to 1999 and later an CM for NEOS. 

	\subsubsection{BZÖ}
	
	Of the eight BZÖ CMs shown in Table ~\ref{tab:ch6_bz_misclassified_mps} seven were labelled as FPÖ and one as ÖVP
	\begin{table*}
    \small
    \centering
    \begin{tabular}{|p{3cm}|l|l|l|l|l|l|l|l|l|l|}
    \hline
        Speaker & SPÖ & ÖVP & FPÖ & GRÜNE & LIF & BZÖ & NEOS & TS & PILZ & predicted \\ \hline
        Ernest Windholz & 0.14 & 0.14 & 0.4 & 0.04 & 0 & 0.22 & 0.01 & 0.04 & 0 & FPÖ \\ \hline
        Gerhard Huber & 0.11 & 0.07 & 0.38 & 0.03 & 0 & 0.34 & 0 & 0.06 & 0 & FPÖ \\ \hline
        Hubert Gorbach & 0.12 & 0.42 & 0.39 & 0.02 & 0 & 0.04 & 0 & 0 & 0 & ÖVP \\ \hline
        Karin Gastinger & 0.18 & 0.32 & 0.46 & 0.01 & 0 & 0.02 & 0 & 0 & 0 & FPÖ \\ \hline
        Peter Westenthaler & 0.08 & 0.12 & 0.4 & 0.09 & 0 & 0.29 & 0.01 & 0.01 & 0 & FPÖ \\ \hline
        Sigisbert Dolinschek & 0.16 & 0.12 & 0.49 & 0.03 & 0 & 0.18 & 0.01 & 0.01 & 0 & FPÖ \\ \hline
        Veit Schalle & 0.2 & 0.16 & 0.38 & 0.04 & 0 & 0.18 & 0.01 & 0.04 & 0 & FPÖ \\ \hline
        Wolfgang Spadiut & 0.15 & 0.1 & 0.49 & 0.04 & 0 & 0.17 & 0.01 & 0.03 & 0 & FPÖ \\ \hline
    \end{tabular}
	\caption{Party affiliation score of misclassified BZÖ members}
	\label{tab:ch6_bz_misclassified_mps}
	\end{table*}

	\subsubsubsection{BZÖ → ÖVP}
	The one BZÖ CM who was misclassified as ÖVP was Hubert Gorbach who was Minister for Transport, Technology and Innovation as well as Vice Chancellor from 2003 to 2007. 
	His time in Parliament overlaps mostly with the time of the BZÖ/FPÖ coalition with ÖVP from 2002 to 2006. 

	\subsubsubsection{BZÖ → FPÖ}

	Five of the seven BZÖ CMs wrongly labelled as FPÖ had actually switched parties from FPÖ to BZÖ.
	The CMs with an actual FPÖ background are: Ernest Windholz, Peter Westenthaler, Sigisbert Dolinschek, Wolfgang Spadiut and Christoph Hagen.

	Christoph Hagen switched party many times. 
	He was an CM for FPÖ until 2004 when he joined BZÖ.
	From 2008 to 2012 he became an CM for BZÖ again and left the party to be independent for a short time.
	After less than a month of being independent he became an CM for STRONACH from 2012 to 2017.
	Finally, he left STRONACH and became independent again for just three months before he left Parliament.

	Gerhard Huber was CM for BZÖ from 2008 to 2009, became an independent CM from 2009 to 2010 and then switched back to BZÖ from 2010 to 2013. 
	He was asked to leave the Parliamentary Group for some of that year because of a case against him.

	Karin Gastinger was a Minister of Justice from 2004 to 2007.
	She was never a member of a party, but was nominated by the BZÖ.
	Veit Schalle was an CM from 2006 to 2008 when he was already retired.
	Before he worked for the REWE group and was active against "Kärntner Slowenen".
	He has connections to right wing activism in Carinthia.
	There are quotes about the Third Reich's market politics in media coverage.

	\subsubsection{NEOS}
	The five NEOS CMs who were misclassified, three as FPÖ, one as SPÖ and one as GRÜNE, are shown in Table ~\ref{tab:ch6_neos_misclassified_mps}

	\begin{table*}
    \small
    \centering
    \begin{tabular}{|p{3cm}|l|l|l|l|l|l|l|l|l|l|}
    \hline
        Speaker & SPÖ & ÖVP & FPÖ & GRÜNE & LIF & BZÖ & NEOS & TS & PILZ & predicted \\ \hline
        Christoph Vavrik & 0.16 & 0.16 & 0.21 & 0.17 & 0.01 & 0.05 & 0.13 & 0.06 & 0.04 & FPÖ \\ \hline
        Gerald Loacker & 0.18 & 0.11 & 0.27 & 0.12 & 0.01 & 0.09 & 0.16 & 0.05 & 0.02 & FPÖ \\ \hline
        Irmgard Griss & 0.28 & 0.14 & 0.13 & 0.24 & 0.02 & 0.05 & 0.09 & 0.01 & 0.06 & SPÖ \\ \hline
        Josef Schellhorn & 0.1 & 0.15 & 0.32 & 0.08 & 0.01 & 0.05 & 0.17 & 0.1 & 0.02 & FPÖ \\ \hline
        Nikolaus Scherak & 0.15 & 0.06 & 0.24 & 0.24 & 0 & 0.07 & 0.19 & 0.01 & 0.04 & GRÜNE \\ \hline
    \end{tabular}
	\caption{Party affiliation score of misclassified NEOS members}
	\label{tab:ch6_neos_misclassified_mps}
	\end{table*}

	\subsubsubsection{NEOS → SPÖ}
	Irmgard Griss was an CM for NEOS from 2017 to 2019 even though she was actually never a party member of NEOS.
	She is a judge and her speech predictions are somewhere between SPÖ/ÖVP and NEOS. 
	\subsubsection{NEOS → FPÖ}
	Christoph Vavrik was an CM for NEOS from 2013 to 2017 but joined the ÖVP Parliamentary Group for six months in 2017.
	He was banned from the Parliamentary Group of NEOS because he was accused of making a homophobic post on Facebook, and ÖVP took him in.
	Gerald Loacker was an CM for NEOS from 2013 to 2024 and active in the fraternity KaV Norica Wien.
	Before his time in Parliament as a NEOS CM, he was an ÖVP member.
	Josef Schellhorn was a member of ÖVP until 2013.
	He then switched to become an CM for NEOS from 2014 to 2021 and also from 2024 to 2025.
	From 2024 onwards he is Minister for Outer Agendas and European Policies.

	\subsubsubsection{NEOS → GRÜNE}
	Nikolaus Scherak was an CM for NEOS from 2013 to 2014. 

	\subsubsection{STRONACH}

	As detailed in Table ~\ref{tab:ch6_stronach_misclassified_mps}, nine STRONACH CMs were misclassified, six as FPÖ, two as BZÖ and one as ÖVP.
	\begin{table*}
    \centering
    \small
    \begin{tabular}{|p{3cm}|l|l|l|l|l|l|l|l|l|l|}
    \hline
        Speaker & SPÖ & ÖVP & FPÖ & GRÜNE & LIF & BZÖ & NEOS & TS & PILZ & predicted \\ \hline
        Elisabeth \mbox{Kaufmann-Bruckberger} & 0.17 & 0.16 & 0.43 & 0.01 & 0 & 0.18 & 0 & 0.05 & 0 & FPÖ \\ \hline
        Erich Tadler & 0.2 & 0.12 & 0.23 & 0.02 & 0 & 0.26 & 0 & 0.16 & 0 & BZÖ \\ \hline
        Georg Vetter & 0.14 & 0.26 & 0.31 & 0.08 & 0.01 & 0.06 & 0.05 & 0.1 & 0.01 & FPÖ \\ \hline
        Leopold Steinbichler & 0.2 & 0.2 & 0.33 & 0.05 & 0 & 0.07 & 0.04 & 0.1 & 0.01 & FPÖ \\ \hline
        Martina Schenk & 0.11 & 0.06 & 0.24 & 0.05 & 0 & 0.33 & 0.02 & 0.17 & 0.01 & BZÖ \\ \hline
        Robert Lugar & 0.05 & 0.04 & 0.3 & 0.05 & 0 & 0.29 & 0.03 & 0.23 & 0.01 & FPÖ \\ \hline
        Rouven Ertlschweiger & 0.21 & 0.32 & 0.23 & 0.04 & 0.01 & 0.02 & 0.05 & 0.11 & 0.01 & ÖVP \\ \hline
        Ulrike Weigerstorfer & 0.18 & 0.09 & 0.35 & 0.13 & 0 & 0.03 & 0.04 & 0.16 & 0.01 & FPÖ \\ \hline
        Waltraud Dietrich & 0.11 & 0.19 & 0.41 & 0.02 & 0 & 0.11 & 0.02 & 0.15 & 0 & FPÖ \\ \hline
    \end{tabular}
	\caption{Party affiliation score of misclassified STRONACH members}
	\label{tab:ch6_stronach_misclassified_mps}
	\end{table*}

	\subsubsubsection{STRONACH → ÖVP}
	Whilst Rouven Ertlschweiger was  an CM for STRONACH from 2014 to 2015, he then switched from STRONACH and became an ÖVP CM from 2015 to 2017.

	\subsubsubsection{STRONACH → FPÖ}

	Elisabeth Kaufmann-Bruckberger was an CM for BZÖ from 2011 to 2012 and was briefly independent before she joined the STRONACH Parliamentary Group from 2012 to 2013. 
	She was an FPÖ member before she joined BZÖ and was politically active on a local political level before and after her time in Parliament. 

	Leopold Steinbichler was an CM for ÖVP from 1997 to 2003 and reentered Parliament as an CM for STRONACH from 2013 to 2017.
	
	Two of the six misclassified CMs switched parties from STRONACH to ÖVP, but nonetheless were labelled as FPÖ members according to their speeches:
	Georg Vetter and Marcus Franz were both CMs for STRONACH from 2013 to 2015 and switched to being CMs for ÖVP in 2015 after they were invited by Reinold Lopatka to join ÖVP Parliamentary Group. 
	Franz had to leave, due to a critical statement he made regarding Angelika Merkel and her policies in 2016 and he became independent after that. 
	Robert Lugar was an CM who switched parties multiple times. 
	He was an CM for BZÖ from 2008 to 2011, an independent CM after that and an CM for STRONACH from 2012 to 2017.
	He then joined the FPÖ Parliamentary Group from 2017 to 2019. 
	So to a degree the FPÖ prediction is not completely wrong. 
	Ulrike Weigerstrofer was an CM for STRONACH from 2013 to 2017.
	She worked for Stronach's Magna Group.

	Waltraud Dietrich's prediction is also partially correct.
	She was an CM for STRONACH from 2013 to 2017, but before 2012 was an FPÖ member. 

	\subsubsubsection{STRONACH → BZÖ}

	Erich Tadler was rightfully predicted as BZÖ. From 2008 to 2010 he was actually a BZÖ CM.
	However, following a period as an independent CM from 2010 to 2012, he then joined STRONACH as an CM from 2012 to 2013. 
	Likewise, Martina Schenk was an CM for both BZÖ from 2008 to 2013 and then STRONACH from 2013 to 2017.

	\subsubsection{PILZ}
	Table ~\ref{tab:ch6_pilz_misclassified_mps}: There were only six CMs in the entire data set period.
	None of the six were correctly classified as being a PILZ CM.
	One explanation could be that the roughly 90 speeches were not enough labelled data for training a text classifier for PILZ.
	Five of the PILZ CMs were misclassified as SPÖ. 
	The other, Alfred Noll, was misclassified as GRÜNE.
	One of the five, Alma Zadic, was an  CM for PILZ from 2017 to 2019, an independent CM from July to October 2019 and then ran for GRÜNE in the following elections and served as a GRÜNE CM from 2019 to 2020.
	Later, from 2020 to 2025, she was Minister of Justice for GRÜNE.
	\begin{table*}
    \centering
    \small
    \begin{tabular}{|p{3cm}|l|l|l|l|l|l|l|l|l|l|}
    \hline
        Speaker & SPÖ & ÖVP & FPÖ & GRÜNE & LIF & BZÖ & NEOS & TS & PILZ & predicted \\ \hline
        Alfred J. Noll & 0.17 & 0.1 & 0.24 & 0.28 & 0.01 & 0.03 & 0.13 & 0.02 & 0.04 & GRÜNE \\ \hline
        Alma Zadic & 0.23 & 0.14 & 0.11 & 0.15 & 0 & 0.03 & 0.17 & 0.05 & 0.12 & SPÖ \\ \hline
        Daniela \mbox{Holzinger-Vogtenhuber} & 0.41 & 0.1 & 0.08 & 0.16 & 0 & 0.02 & 0.09 & 0.08 & 0.04 & SPÖ \\ \hline
        Martha Bißmann & 0.28 & 0.14 & 0.09 & 0.13 & 0 & 0.03 & 0.13 & 0.02 & 0.17 & SPÖ \\ \hline
        Peter Kolba & 0.37 & 0.17 & 0.12 & 0.09 & 0 & 0.03 & 0.07 & 0.07 & 0.1 & SPÖ \\ \hline
        Stephanie Cox & 0.31 & 0.11 & 0.09 & 0.2 & 0 & 0.01 & 0.18 & 0.04 & 0.05 & SPÖ \\ \hline
    \end{tabular}
	\caption{Party affiliation score of misclassified PILZ members}
	\label{tab:ch6_pilz_misclassified_mps}
	\end{table*}

	Daniela Holzinger-Vogtenhuber was an CM for SPÖ from 2013 to 2017 before she became an independent CM from July 2017 to November 2017 and then switched to PILZ where she was an CM from	 2017 to 2019. 

	Martha Bißmann was an CM for PILZ from 2017 to 2018. 
	She was active in the green student organisation but later went campaigning for Irmgard Griess at the presidential elections. Irmgard Griess went to NEOS. 

	Peter Kolba was an CM for PILZ from 2017 to 2018.
	When he left Parliament, Peter Pilz took over his seat.
	Stephanie Cox was an CM for PILZ from 2017 to 2019.
	Her speeches are quite spread between SPÖ with 31\%, 20\% for GRÜNE and 18\% for NEOS.
	
The first step of our approach is to train a classifier. The goal of the trained model is to give a the prediction of the party affiliation based on text. Given a speech,  should return a probability distribution across all parties in Parliament. 

\subsection{training}
Because the aim is to collect the prediction for every speech, a leave-one-fold-out (n=5) training architecture was chosen. Five BERT models will be trained on 80\% of the data, leaving out 20\% each.
	
\subsubsection{Training folds}
To ensure that the models are somewhat comparable, each fold needed roughly the same distribution of training data.
The data was divided into five folds trying to create folds that have a similar distribution of parties compared to the data set as a whole. 
Additionally, we wanted to be able to predict each speaker's party affiliation. 
In order to be able to get a prediction, all CMs needed a model where their speeches were not included in the training data. 
So every CM is not in the training data of one of the five models.
	
First we sorted the speeches into speakers, and then we ranked the CMs with the number of speeches.
Then the CMs and their speeches were looped over for each party and distributed. 
One by one the speeches of the CMs were distributed amongst the five folds. 
We ended up with folds with a distribution similar to Figure ~\ref{fig:ch3_mps}: 

For each model one fold was kept as the testing set. 
The other four folds were used to train the models.
As one can see the smaller parties are more difficult to share between five, nevertheless each fold has every party.
	
\subsubsection{Parameters}

The hugging face PyTorch transformers library~{https://pytorch.org/} was used to set up the training. 
In this paper we fine-tuned the \texttt{bert-base-german-cased} \href{https://huggingface.co/google-bert/bert-base-german-cased}{pretrained Model}.
Choosing the context length is a balance between efficiency and performance and for this case the length of 256 tokens was sufficient.

The standard deviation of words lies around 528 words.
Half of the speeches are under 300 words, so using a 512 token context window size would result in even more empty spaces.
Roughly a quarter of speeches are between 300 and 600 words long.
Because splitting is inevitable for those and the longer speeches, it is more convenient to split the speeches into smaller chunk sizes.
That should give more balanced chunks than having a big chunk and a mini one. 
For every model we split each speech into chunks with n=256 tokens.
Our final training dataset consists of speech-chunks, their party label as well as an ID to aggregate the chunks back to speech level.
	
\subsubsection{Model architecture}
	
We fed the chunks to a BERT Classifier with a BERT layer, the pretrained model and a drop out layer  with 10\% drop out for regularisation.
Finally, a linear layer with output classes for each party (n=9) to perform the classification.

We used a batch size of 20 and two training epochs. 
With a smaller test sample more epochs were tried, but there were no improvements in performance.
The BertTokenizer was used for tokenization.
The training loop passes a batch, the gradients are reset and the input is sent to CPU, then the attention\_mask does the padding. 
The labels are sent to CPU too. 
In the forward pass the models logits are calculated and taken into the CrossEntropyLoss function. 
The loss is being backpropagated with an optimizer step and then returned. 

The probability distribution for each chunk in the evaluation was collected and the chunk predictions were aggregated on speech level (creating a simple mean).
The aggregated probability distribution of the speeches was linked back with the data set.
The variable "predicted\_party" contains the max probability party after taking the mean of all chunks.
We did this for each of our five trained models and joined them back into the dataset. 
Later, we also tested the 1,555 speeches labelled as independent on all five models and took the mean of the predictions for each model to add to the final data set.

\end{document}

%% file: classificationmodelshort.tex
We train a BERT-based classifier to predict party affiliation from parliamentary speeches, outputting a probability distribution over all parties. BERT was selected for its ability to capture context, syntax, and phrases, critical for distinguishing party-specific language patterns. In preliminary experiments, BERT considerably outperformed logistic regression, expectedly confirming the importance of contextual modelling for this task. We compare the results in section~\ref{sub:res_table}. We also briefly considered using prompted generative LLMs, but found that for our purposes BERT provided adequate performance at a fraction of the computational cost.

%% file: switchingaffiliationshort.tex
\paragraph{Switching affiliation}
We first investigated whether party switching could be predicted from speech patterns. Classifier performance on this task was poor. Our analysis shows that CMs typically adapt their speech behaviour to their new party after switching, not before. Most switches involved newly created parties rather than established ones. For example, Josef Bucher, a long-serving FPÖ politician who switched around BZÖ’s creation date, shows a clear shift in predicted probabilities, with the orange line (BZÖ) rising and the blue line (FPÖ) falling (Figure~\ref{fig:ch6_josef_bucher}), a pattern observed across multiple cases (Appendix).

Possible explanations include:       (1) newly founded parties may require time to develop distinct linguistic identities; (2) some parties exhibit less internal stylistic variation, obscuring switches; or (3) linguistic adaptation may inherently follow party changes. Regardless, results confirm that party-specific speech patterns are discernible, as evidenced by clear clustering in our t-SNE visualisation.

%% file: conclusionshort.tex
\section{Conclusion}
From our analysis, based on a dataset of Austrian parliamentary speeches created as part of this work, we can draw the following conclusions. (1) The model captures party political affiliations using parliamentary speeches. The Visualisation of all CMs in the t-SNE plot from Figure~\ref{fig:ch5_tsne_mps} can provide an affirmative answer to our research question Q1.
To answer Q2, firstly we see that in the trend in Figure ~\ref{fig:ch6_misclassified_sp} and Figure ~\ref{fig:ch6_misclassified_vp} coalition partners exhibit increasing linguistic similarity and also parties in opposition diverge linguistically from governing parties as seen in Figure ~\ref{fig:ch6_misclassified_fp}. Regarding Q3, CMs typically adapt their speech behaviour after switching parties. Figures~\ref{fig:ch6_misclassified_sp} to \ref{fig:ch6_misclassified_fp} support the first two hypotheses, showing that misclassifications often occur toward coalition partners. For the third, most adaptation happens after a switch, with some exceptions (please see the appendix for more elaborate individual-level case studies).

The model effectively predicts party affiliation and captures political proximity, but does not provide feature-level explanations. 
Its strength lies in aggregating large volumes of speeches to empirically identify patterns, particularly in temporal or aggregated analyses. We make all model code and outputs publicly available to facilitate further research on this dataset and topic. Future work could include a straightforward extension of our analysis to German data by testing with the already fine-tuned models or applying the training and evaluation method to other EU countries and languages, which are partially available in the ParlSpeech V2 dataset or to other similar datasets. Moreover, a particularly exciting avenue of future work would be to explore connections between speech behaviour and policy outcomes or public opinion surveys.

%% file: limitationsshort.tex
\section{Limitations}

The model performs poorly at the individual speech level, likely because speeches often address specific topics with limited input. Performance improves at the aggregated CM level, where topical differences average out. However, the model remains a black box: we explored feature-based explanation methods (SHAP \cite{lundberg2017unifiedapproachinterpretingmodel} and DIG \cite{sanyal2021discretizedintegratedgradientsexplaining}) outside the main scope of this work, but they failed to provide meaningful insights. A more detailed discourse-level analysis, such as typically done manually in the social sciences is not supported by our approach, since it does not reveal whether predictions rely on lexical features (e.g., numbers, questions), stylistic variation, or other factors, a limitation typical of neural models.
While this study provides considerable empirical conclusions, additional qualitative analysis is needed to further the insights. 

\section{Acknowledgements}

We would like to thank the anonymous reviewers for their helpful comments and suggestions. This research has been funded by the Vienna Science and Technology Fund
(WWTF)[10.47379/VRG19008] "Knowledge-infused Deep Learning for Natural Language Processing", the Horizon Europe (European Commission), Marie Skłodowska-Curie Action  postdoc grant (MSCA)[101212164] "Active Learning with Large Language Models for Cost Efficient Pseudo-labeling in Computational Social Science", as well as CLARIAH-AT. \footnote{https://clariah.at/en/}